\documentclass{article}

\usepackage[preprint,nonatbib]{airov24}

\usepackage[utf8]{inputenc} 
\usepackage[T1]{fontenc}    
\usepackage{hyperref}       
\usepackage{url}            
\usepackage{booktabs}       
\usepackage{amsfonts}       
\usepackage{nicefrac}       
\usepackage{microtype}      
\usepackage{xcolor}         
\usepackage{amsmath}
\usepackage{graphicx}
\usepackage{subcaption}
\usepackage{caption}
\usepackage{makecell}
\usepackage{multirow}
\usepackage{amsmath,amsfonts}

\title{Enhancing Transparent Object Pose Estimation: A Fusion of GDR-Net and Edge Detection}

\author{
  Tessa Pulli \\
  Automation and Control Institute\\
  TU Wien\\
  Vienna, Austria\\
  \texttt{pulli@acin.ac.tuwien.at} \\
  \And
  Peter Hönig \\
  Automation and Control Institute\\
  TU Wien\\
  Vienna, Austria\\
  \texttt{hoenig@acin.ac.tuwien.at} \\
  \AND
  Stefan Thalhammer \\
  Department of Industrial Engineering \\
  University of Applied Sciences Technikum Wien \\
  Vienna, Austria \\
  \texttt{stefan.thalhammer@technikum-wien.at } \\
  \And
  Matthias Hirschmanner \\
  Automation and Control Institute\\
  TU Wien\\
  Vienna, Austria\\
  \texttt{hirschmanner@acin.ac.tuwien.at} \\
  \AND
  Markus Vincze \\
  Automation and Control Institute\\
  TU Wien\\
  Vienna, Austria\\
  \texttt{vincze@acin.ac.tuwien.at} \\
}

\begin{document}

\maketitle

\begin{abstract}
 Object pose estimation of transparent objects remains a challenging task in the field of robot vision due to the immense influence of lighting, background, and reflections. However, the edges of clear objects have the highest contrast, which leads to stable and prominent features. We propose a novel approach by incorporating edge detection in a pre-processing step for the tasks of object detection and object pose estimation. We conducted experiments to investigate the effect of edge detectors on transparent objects. We examine the performance of the state-of-the-art 6D object pose estimation pipeline GDR-Net and the object detector YOLOX when applying different edge detectors as pre-processing steps (i.e., Canny edge detection with and without color information, and holistically-nested edges (HED)). We evaluate the physically-based rendered dataset Trans6D-32 K of transparent objects with parameters proposed by the BOP Challenge. Our results indicate that applying edge detection as a pre-processing enhances performance for certain objects. 
\end{abstract}

\section{Introduction}
\begin{figure*}[htb]  
  \centering
  \includegraphics[width=\textwidth]{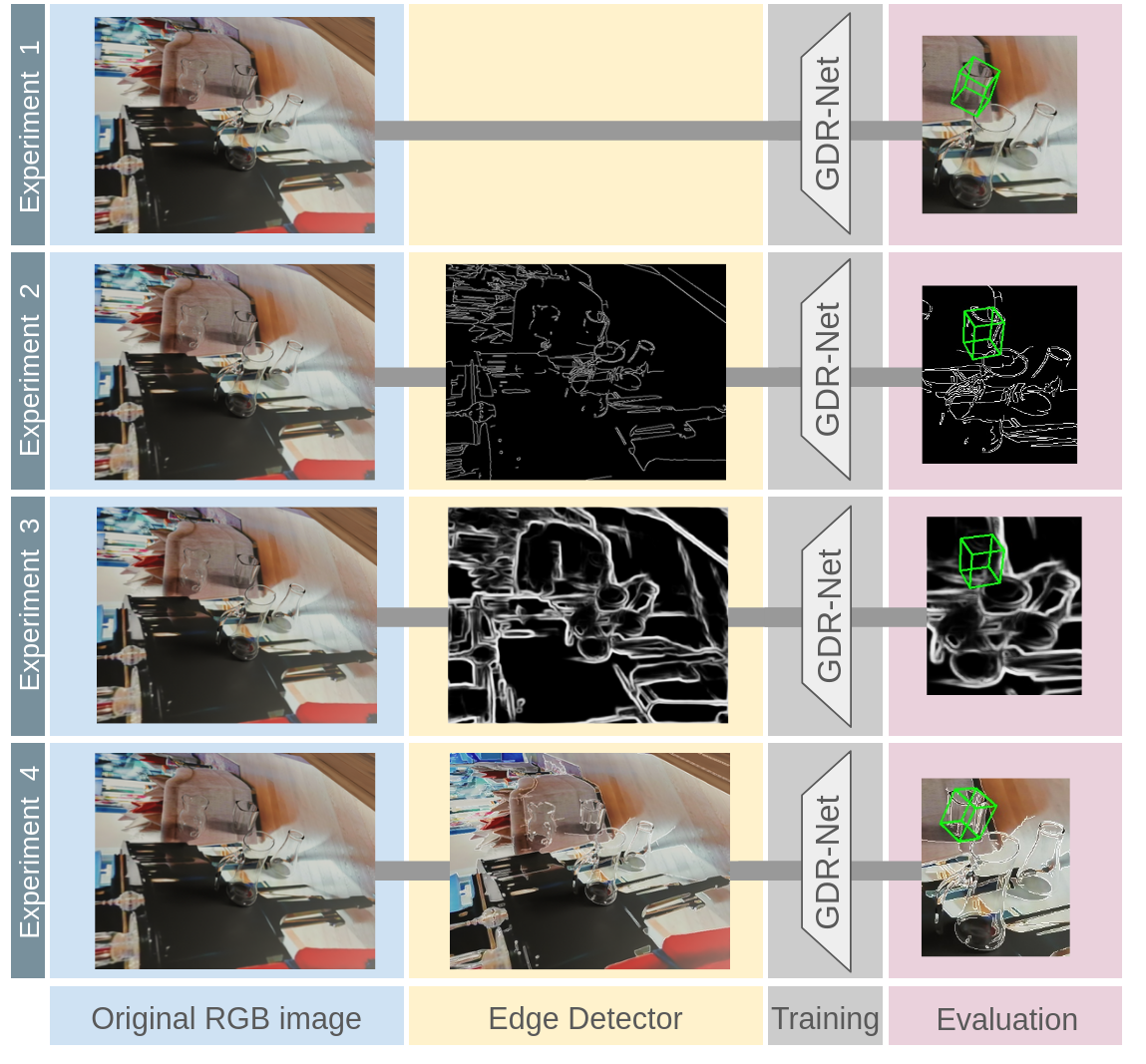}
  \caption{\textbf{Experimental setup.} We conduct four sets of experiments: In the first experiment, the Trans6D-32K dataset without any augmentation (vanilla) is trained and tested. The second experiment investigates the performance when Canny is applied before training. The third experiment involves HED as an pre-processing step. Finally, we conducted a set of experiments incorporating color information and canny edges.}
  \label{fig:setup}
\end{figure*}

6D object pose estimation aims to determine the position and orientation of objects within a 3D space. It has been widely used in real-world robotics applications, such as object grasping \cite{zhu2014single, tremblay2018deep} and manipulation \cite{collet2011moped}. While transparent items are among the most prevalent objects in daily life, their 6D pose estimation remains a challenging task.
Their visual appearance is highly determined by the lighting conditions, backgrounds, and its reflective properties.
Background patterns are reproduced by the object's surface which complicates the extraction of recognizable features.
Even if RGB-D methods can usually achieve higher performances \cite{hinterstoisser2013model, hinterstoisser2016going}, the depth information of transparent objects is noisy and incomplete.
Because depth information does not incorporate valuable information for transparent objects, we rely on RGB images only.
Still, the edges of clear objects have the highest contrast presenting stable and prominent features bearing potential for feature extraction \cite{trans6d}.
Edge detection filters emphasize the outlines of objects in RGB images, enabling the network to obtain more precise information about the transparent object's boundaries.
Several works \cite{er-pose, brad} demonstrate the capabilities of introducing edge detection to CNNs.
As their contribution suggests the effectiveness of this approach, we introduce two types of edge detectors to a state-of-the-art object pose estimation method and investigate the effects on 6D object pose estimation of transparent objects.
\\
Figure \ref{fig:setup} illustrates the methodology of our experiments.
The first set of experiments is conducted on the dataset Trans6D-32 K \cite{trans6d} comprising RGB images, while the second and third set of experiments incorporates an additional pre-processing step performing edge detection with Canny edge detection \cite{canny} and holistically-nested edge detection \cite{xie15hed}. In a fourth set of experiments, we combine RGB information with canny edges \cite{canny}.
Our work focuses on detecting objects and recovering the corresponding 6D pose with challenging features.
We investigate the effect of edge representations by attempting to achieve a more precise object representation by depicting only the outlines of the corresponding objects.
In summary, the paper has the following key contributions:

\begin{itemize}
\item We propose a novel approach addressing the challenge of transparent 6D object pose estimation from RGB images.
We introduce edge filters as a pre-processing step to highlight the object boundaries.
\item We design a set of experiments to demonstrate that edge detection can enhance the performance of an object pose estimation pipeline. The experiments prove that edge detection has the potential to improve 6D object pose estimates of transparent objects.
\end{itemize}

The rest of this work is organized as follows: Section II introduces the related work of 6D object pose estimation on RGB images, 6D object pose estimation on transparent objects, and edge representation; Section III describes our method of the experiments, Section IV presents the experimental setup, evaluation, and results; and Section V concludes the work.

\section{Related Work}
\vspace{-3mm}
In this section, we discuss the related works by revisiting 6D object pose estimation on RGB images, for transparent object, and representation of edges.

\subsection{6D Object Pose Estimation on RGB Images}
When considering the input data, 6D object pose estimation can be divided into RGB-D-based and RGB-based methods.
RGB-D approaches consider 2D RGB data as well as depth information to achieve a pose estimate. 
While these methods achieve high performances on textured data \cite{pvn3d, g2l, densefusion}, errors occur when being applied to transparent objects. 
State-of-the-art sensor systems fail when it comes to obtaining depth data of transparent objects properly. For this reason, we concentrate on methods using only RGB information.
Most current approaches rely on multi-stage networks instead of regressing the pose directly.
To extract intermediate features approaches such as segmentation masks \cite{Hu_2020_CVPR}, key points \cite{kundu2019object}, corner points of bounding boxes \cite{rad2017bb8}, or feature maps \cite{gdrnet} are used. This results in 2D-3D correspondences meaning that points in the 2D image are matched to their position in three-dimensional space \cite{huang1994motion}.
Then, the pose is estimated based on the retrieved features by applying an additional Perspective-n-Point (PnP) algorithm \cite{pavlakos20176, tjaden2017real}. The PnP problem describes the problem of finding the relative pose between an object and a camera from a set of n pairings between 3D points of the object and their corresponding 2D projections on the focal plane \cite{wu2006pnp}. 


E.g. the Geometry Guided Direct Regression Network (GDR-Net) \cite{gdrnet} combines direct and geometry-based indirect approaches, divides objects into fragments and outputs the pose estimate directly. 
The approach decouples 2D object detection and 6D object pose estimation.
The method identifies all objects by means of a detector and identifies a Region of Interest (RoI) for each detection and predicts a geometric feature map of the RoI.
The Patch-PnP algorithm is then employed to directly regress the 6D object pose. 
Even if the methods above achieve considerable results on textured objects, it is not explictily designed for transparent objects and their unique properties.

\subsection{6D Pose Estimation of Transparent Objects}
Transparent objects cause reflections and refractions of light leading to a non-uniform appearance of the surface.
Correspondences between different parts of the object are challenging to establish when the surface of an object does not have the same texture. 
Due to these unique visual properties, their depth information can not be captured by state-of-the-art sensors.
The lack of distinct textures and features in transparent objects requires other strategies to extract the pose information.
\cite{liu2020keypose, chen2022clearpose, zhang2022transnet} propose networks explicitly designed for estimating poses of transparent objects. 
Zhou et al. \cite{zhou2019glassloc} estimate the probability of being transparent for each pixel with a light-field camera.
Sajan et al. \cite{sajjan2020cleargrasp} employ deep CNNs to infer surface normals, transparent surface masks, and occlusion boundaries. 
These outputs are then used to refine depth estimates of the scene. 
Liu et al. \cite{liu2020keypose} propose with Keypose to train a deep neural network on raw stereo images to predict object poses from 3D keypoints.
The mentioned methods can achieve considerable results on transparent objects.
However, these specific methods bear major limitations restricting their usage in real-world applications.
A frequent assumption is the availability of a 3D model of the corresponding object \cite{glassware, transparent_clutter, mobile_platform}, assume object properties, or require specific equipment \cite{zhou2019glassloc}.
Our method considers the usage of a generic 6D object pose estimation pipeline to avoid the prerequisites which come with pipelines explicitly for transparent objects.

\subsection{Edge Representation}
Several studies have integrated edge representation strategies in CNN in order to achieve better performance results. 
Edge representations have proved themselves to be advantageous when handling texture-less objects \cite{er-pose, d2co}.
Zhang et al. \cite{zhang2022eanet} propose an edge-detection network to estimate the 6D object pose. By implementing a shared-weight edge extractor, edge reconstruction and pose estimation are derived simultaneously.
\cite{er-pose} introduced a two-stage 6D pose estimation method for textureless objects.
Instead of utilizing the object mask in current monocular methods, an edge representation for texture-less objects is proposed.
It was found in the experiments, that directly replacing the object mask with the edge representation can bring a performance improvement in two current two-stage pipelines without any modification \cite{er-pose}.
While Canny edge detection is a well-established method to detect contours reliably, the drawback of the method is the reproduction of foreground and background edges to the same extent \cite{canny}.
Holistically Nested Edge Detection (HED) represents a deep learning methodology that employs multi-layered convolutional neural networks to generate images that highlight contours \cite{xie15hed}.
In contrast to conventional edge detection methods, HED is designed to distinguish between background and foreground features. 
Harary et al. \cite{harary2022brad} use an edge-regularized bridge across domains to map features from one domain to another. 
To address the issues of 6D object pose estimation, we conduct experiments with Canny edge detection and HED in order to enhance the performance of state-of-the-art object pose estimation.

\section{Methods}

We assume a given input image \(X \in \mathbb{R}^2\) from which we want to retrieve the 6D object pose described by the rotation \( R \in \mathbb{R}^3 \) and the translation \( t \in \mathbb{R}^3 \) of the object of interest in the camera coordinate frame. The RGB input images are pre-processed with the function \( f \) mapping the RGB image \(X \) to the edge domain with \( f( X ) = X_{e}\). The image \(X_{e} \in \mathbb{R}^2\) is used to estimate the 6D object pose.

\begin{figure*}[h]  
  \centering
  \label{method}
  \includegraphics[width=\textwidth]{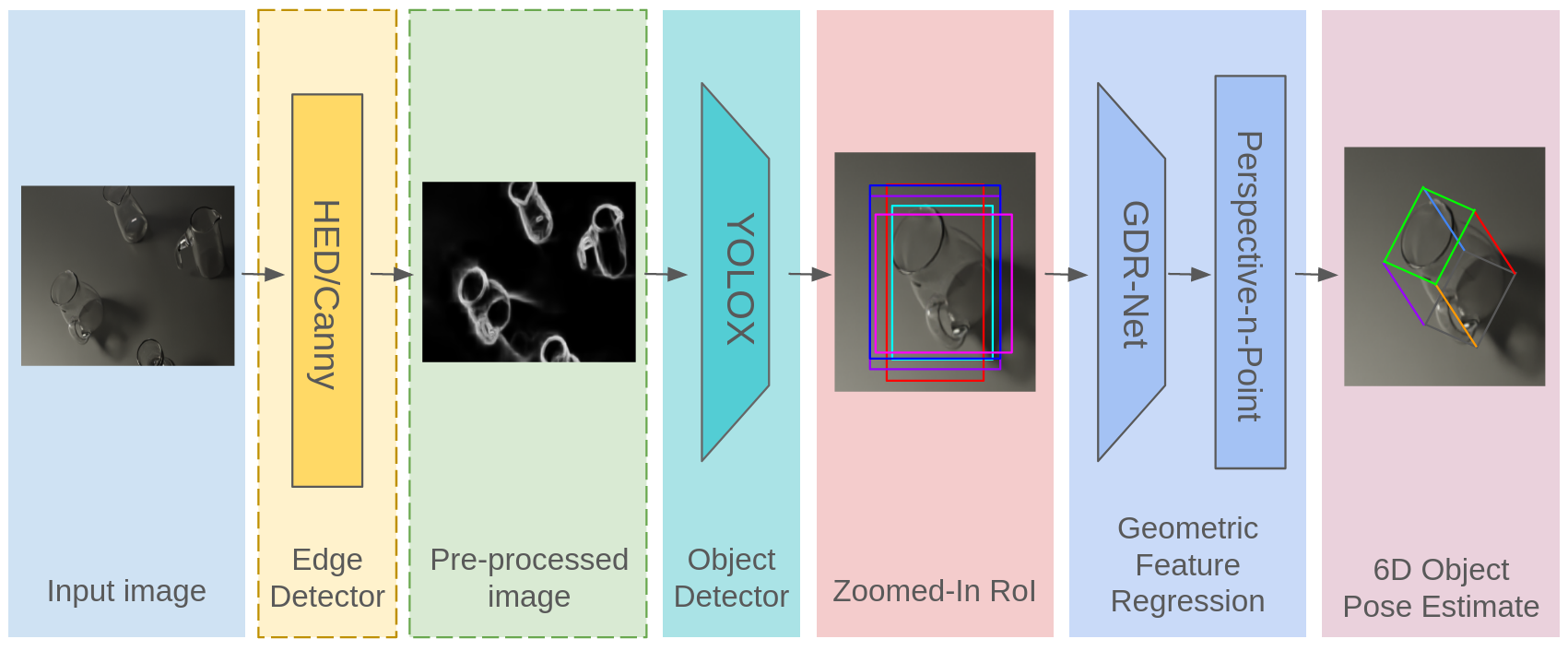}
  \caption{\textbf{Simplified GDR-Net framework and method.} Given an RGB image \(X\), we introduced an optional pre-processing step in which the edge detector is applied and the augmented image is generated. Then, we train the object detector from which zoomed-in RoI are retrieved. This serves as input for the geometric feature regression consisting of the trained GDR-Net and a PnP algorithm to directly regress the 6D object pose estimate.}
  \label{fig:method}
\end{figure*}

Figure \ref{fig:method} presents the schematic overview of the proposed approach. GDR-Net detects the objects of interest using a pre-trained YOLOX object detector. We train the YOLOX detector for each of our datasets to achieve tailored detection results. For each detection, the RoI is retrieved by zooming into the corresponding area. The ROI is then fed to a neural network to predict geometric features from which the geometric pose is then regressed with a PnP step which outputs the 6D pose estimate.

\section{Experiments}
In this section, the experimental setup is discussed. We compare the performance of a RGB datasets on GDR-Net in its original configuration against three augmented configurations. We study the results on the transparent dataset Trans6D-32 K \cite{trans6d}.
We consider for our experiments a two-stage pipeline for which we train the pose estimator integrated in GDR-Net for the 6D object pose and YOLOX for object segmentation. 

\subsection{Implementation details}
 The experiments involving the training of the pose estimator were conducted on an AMD Ryzen 9 5900X CPU with a NVIDIA GeForce RTX 4090 GPU, using the GDR-Net 2022 implementation \cite{gdrnpp}. 
 The Trans6D-32 K \cite{trans6d} dataset was trained for each object class with a batch size of 48 for 100 epochs.
 To evaluate the pose estimation, the publicly available evaluation script of the BOP toolkit is used \cite{bop_toolkit}.
For training the object detector, we used an Intel Core i7-9700K CPU with a Titan V GPU. As suggested by \cite{gdrnpp}, we incorporated YOLOX \cite{ge2021yolox} as the detection method and trained it with data augmentation and ranger optimizer with a batch size of 8. We used the pre-trained YOLOX model provided by \cite{gdrnpp} for the object detection network and evaluated the performance based on the evaluation standards of mainstream object detection models.

\subsection{Dataset}
We evaluate GDR-Net on three configurations of the Trans6D-32 K dataset \cite{trans6d}. Trans6D-32 K is a large-scale synthetic dataset of transparent objects created with Blender. The dataset contains rendered RGB images with different backgrounds, perspectives, and lighting conditions. It contains ten types of transparent household objects of which 5 are symmetrical objects and 5 are asymmetrical objects \cite{trans6d}. We obtained the train-test split for the experiments and selected the same 400 images per object for training as Yu et al. \cite{trans6d} utilized in their experiments because their training in GDR-Net achieved considerable results (mean ADD(-S) of 84.6 over all object classes). Furthermore, 2800 images per class are used as a test set which results in a total of 3200 images per object class. 

\subsection{Pre-processing}
\begin{figure*}[h]  
  \centering
  \includegraphics[width=\textwidth]{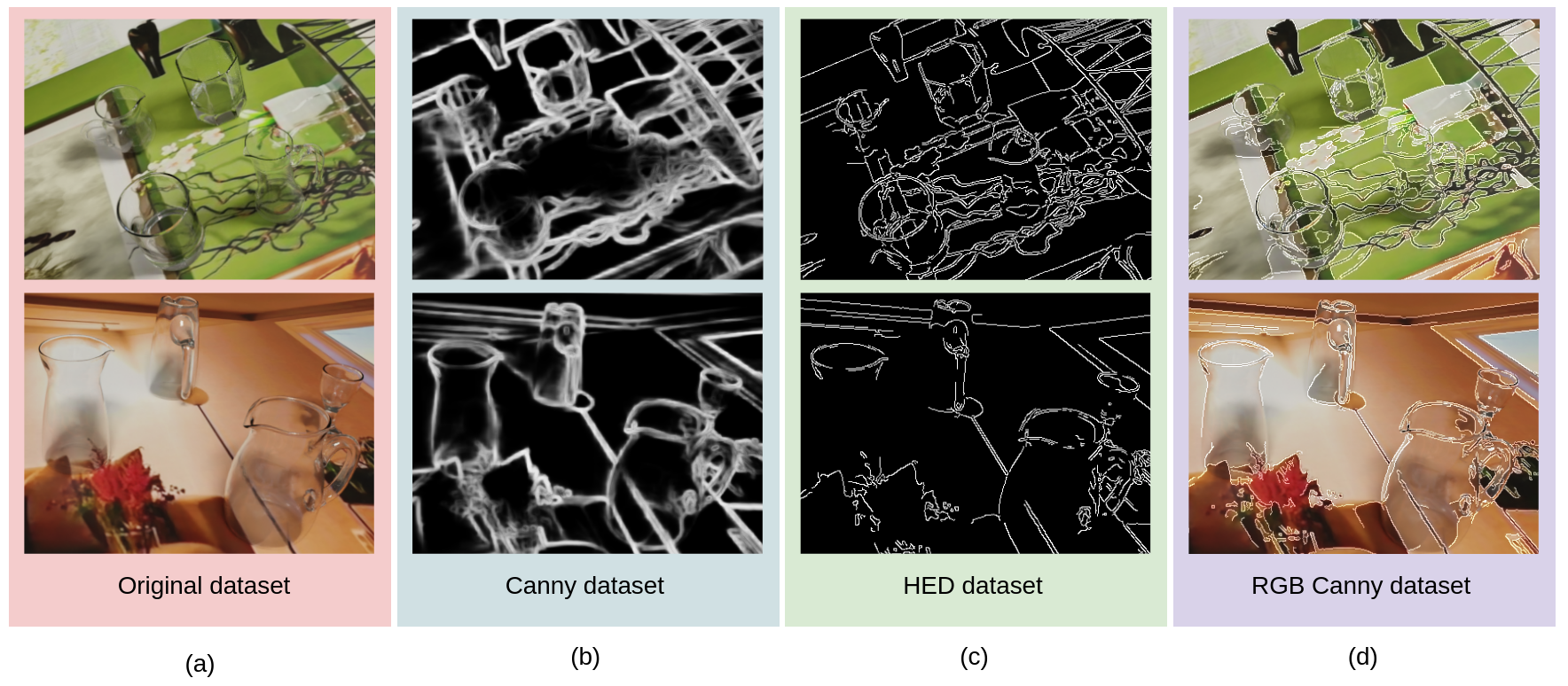}
  \caption{\textbf{Dataset augmentation.} (a) Unaugmented sample images of the Trans6d-32 K dataset (b) Sample images after having applied the Canny edge detector. (c) Sample images after having applied HED. (d) Sample images providing canny edges and color information}
  \label{fig:dataset}
\end{figure*}

For the experiments, three different configurations of the dataset are assessed. Firstly, the original dataset is obtained in its previously described vanilla version. On the second dataset, a Canny filter \cite{canny} is applied. We augmented the images by exploiting the OpenCV Canny function with a lower threshold of 100 and an upper threshold of 200. The third dataset is augmented with the HED \cite{xie15hed}. For that, we employed the method proposed by Xie and Tu \cite{xie15hed} and incorporated the pre-trained models provided on their github. The last dataset incorporates RGB information and canny edges \cite{canny} in one image. In Figure \ref{fig:dataset} samples of each dataset configuration are depicted.

\subsection{Evaluation Metrics}
In the following, we revisit the metrics used for our evaluation. We selected metrics which are present in the bop toolkit \cite{bop_toolkit}.

\subsubsection{Object Pose Estimation}
We evaluate the precision of 6D object pose estimation on the transparent dataset through the average distance ADD metric proposed by Hinterstoisser et al. \cite{add}. 
ADD considers a pose to be accurate if the average distance between the model points in the predicted pose and the ground truth pose is less than 10\% of the model diameter. 
ADD is calculated with a given object with 3D model point set of \(\mathcal{O} = \{x_i \in \mathbb{R}^3 \mid i = 1, 2, \ldots, M\}\) as:

\begin{equation}
\text{ADD} = \frac{1}{\|\mathcal{O}\|} \sum_{x \in \mathcal{O}} \|(R\mathbf{x} + \mathbf{T}) - (R^*\mathbf{x} + \mathbf{T}^*)\|
\end{equation}


where $\left[ R, T \right]$ is the predicted pose and $\left[ R^*, T^* \right]$ is the ground truth pose.

\subsubsection{Object Detection} To evaluate the performance of the detector, we use average precision and average recall. 
Precision measures the accuracy of positive predictions by calculating the ratio of true positive predictions to the sum of true positives and false positives. 
Precision assesses the model's ability to avoid labeling instances as positive when they are actually negative. Precision is defined as:

\begin{equation}
\text{Precision} = \frac{\text{TP}}{\text{TP} + \text{FP}} \quad
\end{equation}

\begin{align*}
\text{where }
& \text{TP} \text{ is the number of correctly predicted positive instances }, \\
& \text{FP} \text{ is the number of instances that were predicted as positive but were actually negative}.
\end{align*}

Recall quantifies the model's capacity to correctly identify all relevant instances in the dataset. It is calculated by dividing the number of true positives by the sum of true positives and false negatives. In essence, recall gauges the model's effectiveness in capturing all instances of a particular class.

\begin{equation}
\text{Recall} = \frac{\text{TP}}{\text{TP} + \text{FN}}
\end{equation}

\begin{align*}
\text{where } \\
& \text{TP} \text{ is the number of correctly predicted positive instances }, \\
& \text{FN} \text{ is the number of actual positive instances that were not predicted as positive}.
\end{align*}

\subsection{Evaluation}
\vspace{-3mm}
We compare our results for the GDR-Net trained 6D object pose estimation and the object detector trained on YOLOX.
We evaluate the methods on each of the introduced datasets.

\begin{table}[h]
  \centering
  \caption{\textbf{Comparison of ADD Recall on the Trans6D-32 K Dataset for each Object Class.} Vanilla refers to the unaugmented dataset. The highest result of each row is printed in \textbf{bold}.}
  \label{gdrn_trans}
  \begin{tabular}{ccccc}
    \toprule
    Object & Vanilla & Canny & HED & RGB Canny\\
    \midrule
     \#1 & \textbf{0.31} & 0.26 & 0.19 & 0.13 \\
     \#2 & 0.27 & 0.25 & 0.26 & \textbf{0.33} \\
     \#3 & 0.20 & \textbf{0.40} & 0.39 & 0.27 \\
     \#4 & \textbf{0.25} & \textbf{0.25} & 0.16 & 0.16 \\
     \#5 & 0.29 & \textbf{0.31} & 0.05 & 0.24 \\
     \#6 & 0.24 & 0.30 & 0.32 & \textbf{0.33} \\
     \#7 & 0.24 & \textbf{0.35} & 0.25 & 0.25 \\
     \#8 & 0.29 & 0.26 & 0.36 & \textbf{0.42} \\
     \#9 & 0.31 & 0.25 & \textbf{0.35} & 0.31 \\
     \#10 & 0.07 & 0.02 & 0.03 & \textbf{0.15} \\
    \midrule
    \textbf{Mean} & 0.25 & \textbf{0.27} & 0.23 & 0.26 \\
    \bottomrule
  \end{tabular}
\end{table}

\subsubsection{Object Pose Estimation Results}
Table~\ref{gdrn_trans} presents the 6D object pose estimation results of the transparent Trans6D-32 K dataset with the augmented datasets (Canny and HED) on GDR-Net. The network is trained and tested for each object of the dataset. The augmented approaches outperform the original dataset in eight out of ten cases. 
While HED achieves the highest ADD(-S) for one object class, the canny approaches (with and without color) give the most accurate ADD(-S) for four object classes each. 
When considering the mean ADD of all objects, the Canny edge augmentation outperforms the three other datasets with an ADD(-S) of 0.27. 

\begin{table}[h]
  \centering
  \caption{\textbf{Comparison of Precision and Recall on the Trans6D-32 K Dataset for each Object Class.} Vanilla refers to the unaugmented dataset. The highest result of each row is printed in \textbf{bold}.}
  \label{yolox_trans}
  \begin{tabular}{ccccc|cccc}
    \toprule
    \multicolumn{5}{c|}{\textbf{Precision (\%)}} & \multicolumn{4}{c}{\textbf{Recall (\%)}} \\
    \addlinespace
    Object & Vanilla & Canny & HED & RGB Canny & Vanilla & Canny & HED & RGB Canny\\
    \midrule
    \#1 & 8.7 & \textbf{25.9} & 25.8 & 6.9 & 15.9 & 22.3 & \textbf{22.3} & 16.3 \\
    \#2 & \textbf{48.7} & 36.4 & 36.3 & 38.7 & 31.1 & 30.7 & 30.7 & \textbf{31.2} \\
    \#3 & \textbf{39.4} & 35.8 & 35.7 & 25.5 & 30.6 & 32.3 & \textbf{32.4} & 28.4 \\
    \#4 & \textbf{36.6} & 26.7 & 26.8 & 33.0 & \textbf{30.4} & 27.9 & 28.0 & 28.8 \\
    \#5 & 26.8 & 29.0 & 29.0 & \textbf{33.1} & 27.8 & 30.6 & \textbf{30.6} & 29.6 \\
    \#6 & \textbf{68.8} & 40.3 & 40.3 & 25.9 & \textbf{34.9} & 32.8 & 32.8 & 25.8\\
    \#7 & \textbf{16.4} & 1.3 & 1.3 & 12.4 & \textbf{22.7} & 0.5 & 0.5 & 19.5\\
    \#8 & 25.2 & 34.8 & 34.7 & \textbf{37.8} & 29.5 & 34.0 & 34.0 & \textbf{34.1}\\
    \#9 & 41.8 & \textbf{45.7} & 45.6 & 45.5 & 34.3 & \textbf{38.5} & 38.4 & 37.1\\
    \#10 & 2.0 & \textbf{5.4} & \textbf{5.4} & 2.0 & 8.1 & 12.8 & \textbf{12.8} & 7.6\\
    \midrule
     Average & \textbf{31.4} & 28.1 & 28.1 & 26.1 & \textbf{26.5} & 26.2 & 26.2 & 25.8\\
    \bottomrule
  \end{tabular}
\end{table}

\subsubsection{Object Detection Results}
Table~\ref{yolox_trans} presents the detection results of the transparent Trans6D-32 K dataset and the augmented datasets (Canny with/without RGB and HED) on YOLOX.
While a positive effect was observable for the 6D object pose estimation when introducing the Canny edge, the unaugmented dataset outperforms both of the datasets depicting only edges.
The additional edge information does not lead to a higher performance in the context of object detection.
The overall poor precision and recall of the dataset can be explained by the small training sample of 400 images per object. 
By increasing the number of training images, it can be expected that a higher performance of the object detector can be achieved.
Furthermore, \cite{trans6d} argue that five of the objects are asymmetrical, while we interpreted those objects to have discrete symmetries.

\section{Conclusion} 
In this work, we investigate the capabilities of edge detection in the context of object detection and pose estimation. 
The key idea is to enhance geometric features by applying edge detection to a dataset before training a CNN with the corresponding data. 
The results of the 6D object pose estimation on Trans6D-32 K indicate that previously applied edge detection can enhance the performance of the CNN. 
Nevertheless, our experiments showed that the introduction of an edge detector to YOLOX does not have a positive effect on the performance. 
We presume, that previously applied edge detection can boost the performance of pose estimators when being applied thoughtfully. 
In the future, we want to extend our work to other challenging scenarios to understand the capabilities of edge detection. 
We consider the nature of the dataset as a major limitation as relatively few training images are used compared to the number of test images. 
Furthermore, the underlying dataset comprised exclusively of synthetic data leading to a purer dataset compared to real-world data. 
For this reason, we intend to conduct further experiments on different datasets.
Additionally, we plan to investigate if similar performance improvements can be achieved in textured objects and consider symmetric properties when evaluating the detection results.

\section*{Acknowledgement}
We gratefully acknowledge the support of the EU-program EC Horizon 2020 for Research and Innovation under project No. I 6114, project iChores.

\end{document}